\documentclass[letterpaper, 10 pt, conference]{ieeeconf}  

\IEEEoverridecommandlockouts                              

\makeatletter
\let\NAT@parse\undefined
\makeatother

\usepackage[nospace]{cite}
\usepackage{amsmath,amssymb,amsfonts}
\usepackage{algorithmic}
\usepackage{graphicx}
\usepackage{textcomp}
\usepackage[dvipsnames]{xcolor}
\usepackage{multirow}
\usepackage{makecell}
\usepackage{tikz}
\usepackage{color, colortbl}
\usepackage{xcolor}
\usepackage{url}
\usepackage[flushleft]{threeparttable}
\usepackage{booktabs}
\usetikzlibrary{arrows.meta}
\let\labelindent\relax
\usepackage{enumitem}
\definecolor{LightBlue}{RGB}{212, 250, 252}
\usepackage{caption}
\usepackage{subcaption}
\usepackage{tabulary}
\usepackage{float}
\usepackage{flushend} 

\usepackage[hidelinks]{hyperref}

\definecolor{Pistachio}{RGB}{140, 212, 126} 
\definecolor{CrayolaYellow}{RGB}{248, 214, 109}
\definecolor{PastelRed}{RGB}{255, 105, 97}
\definecolor{PastelOrange}{RGB}{255, 181, 76}
\definecolor{SoftCharcoal}{RGB}{66,66, 66}

\newcommand{\up}{\,\raisebox{.15em}{\scriptsize$\uparrow$}}
\newcommand{\down}{\,\raisebox{.15em}{\scriptsize$\downarrow$}}

\title{\LARGE \bf 2.5D Object Detection for Intelligent Roadside Infrastructure}

\author{Nikolai Polley$^{\ast1}$, Yacin Boualili$^{\ast1}$, Ferdinand Mütsch$^{1}$, Maximilian Zipfl$^{2}$, \\Tobias Fleck$^{1,2}$, and J. Marius Zöllner$^{1,2}$
\thanks{$^\ast$ Equal contribution.}%
\thanks{$^{1}$ Karlsruhe Institute of Technology (KIT), Research Group Applied Technical-Cognitive Systems, Kaiserstr. 12, Karlsruhe, Germany. {\tt\small \{prename.surname\}@kit.edu}.}
\thanks{$^{2}$ FZI Research Center for Information Technology, \mbox{Technical} Cognitive Systems, Haid-und-Neu Str. 10-14, Karlsruhe, \mbox{Germany}.
	{\tt\small \{surname\}@fzi.de}.}%
}

\usepackage{eso-pic}
\newcommand{\mycopyrighttext}{%
  \footnotesize
  \noindent
  \textcopyright~2025 IEEE. Personal use of this material is permitted.
  Permission from IEEE must be obtained for all other uses, in any current
  or future media, including reprinting/republishing this material for
  advertising or promotional purposes, creating new collective works,
  for resale or redistribution to servers or lists, or reuse of any
  copyrighted component of this work in other works.\\
  2025 IEEE 28th International Conference on Intelligent Transportation Systems (ITSC) - 18-21 November, 2025.
}

\AtBeginDocument{%
  \AddToShipoutPicture*{%
    \AtPageUpperLeft{%
      \put(55, -40){
        \parbox{\dimexpr\textwidth-4pt\relax}{\raggedright \mycopyrighttext}
      }%
    }
  }
}

\begin{document}

\maketitle
\thispagestyle{empty}
\pagestyle{empty}

\begin{abstract} On-board sensors of autonomous vehicles can be obstructed, occluded, or limited by restricted fields of view, complicating downstream driving decisions. Intelligent roadside infrastructure perception systems, installed at elevated vantage points, can provide wide, unobstructed intersection coverage, supplying a complementary information stream to autonomous vehicles via vehicle-to-everything (V2X) communication. However, conventional 3D object-detection algorithms struggle to generalize to the domain shift introduced by top-down perspectives and steep camera angles. We introduce a 2.5D object detection framework, tailored specifically for roadside infrastructure cameras. Unlike conventional 2D or 3D object detection, we employ a prediction approach to detect ground planes of vehicles as parallelograms in the image frame. The parallelogram preserves the planar position, size, and orientation of objects while omitting their height, which is unnecessary for most downstream applications. For training, a mix of real-world and synthetically generated scenes is leveraged. We evaluate generalizability on a held-out camera viewpoint and on adverse-weather scenarios absent from the training set. Our results show high detection accuracy, strong cross-viewpoint generalization, and robustness to diverse lighting and weather conditions. Model weights and inference code are provided at: \\ \url{https://gitlab.kit.edu/kit/aifb/ATKS/public/digit4taf/2.5d-object-detection}
\end{abstract}

\section{Introduction}

\begin{figure}[t]
\centering

\begin{subfigure}[t]{1\linewidth}
  \includegraphics[width=1.0\textwidth]{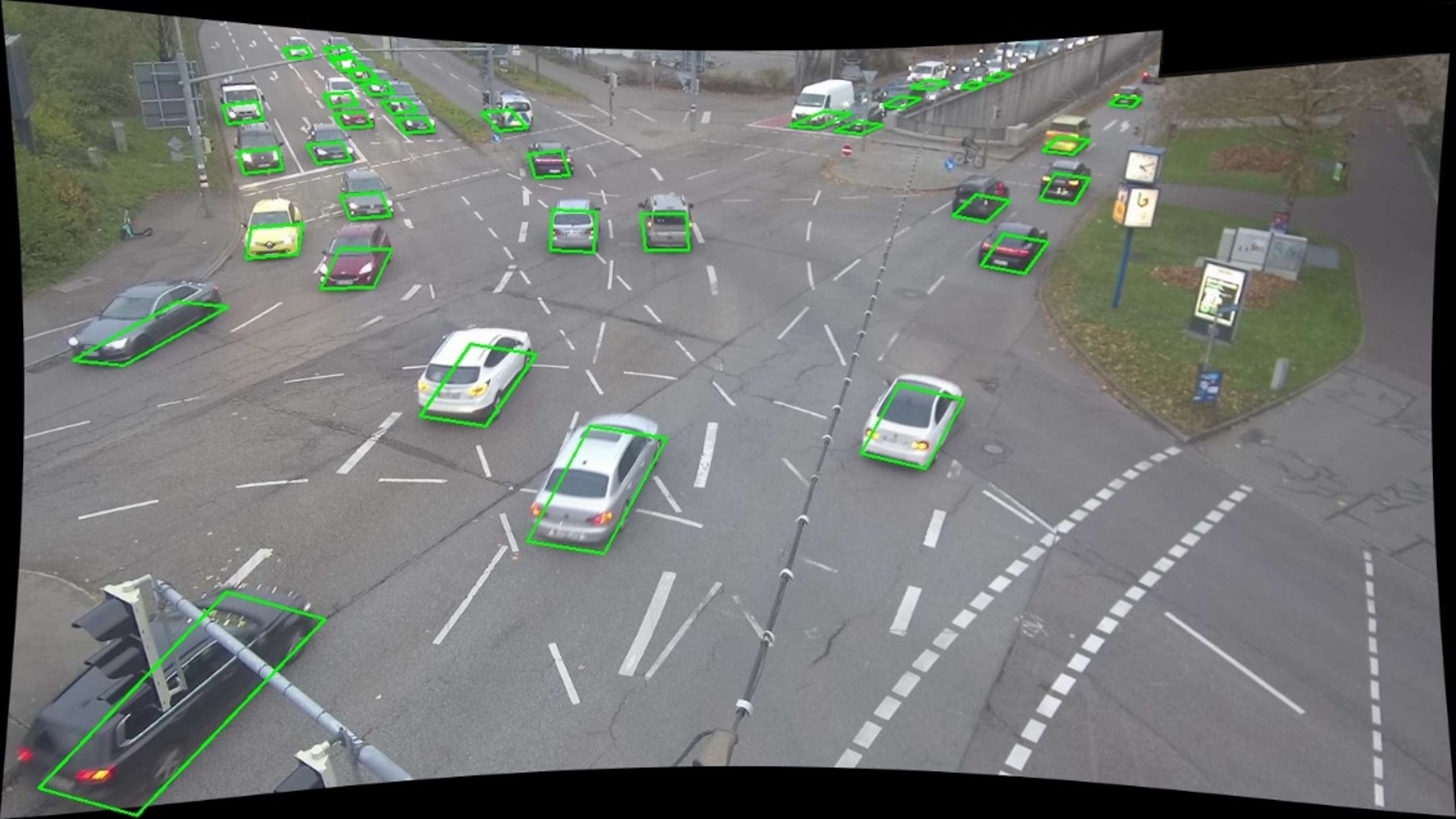}
\end{subfigure}

\vspace{0.05cm}

\begin{subfigure}[t]{1\linewidth}
  \includegraphics[width=1.0\textwidth]{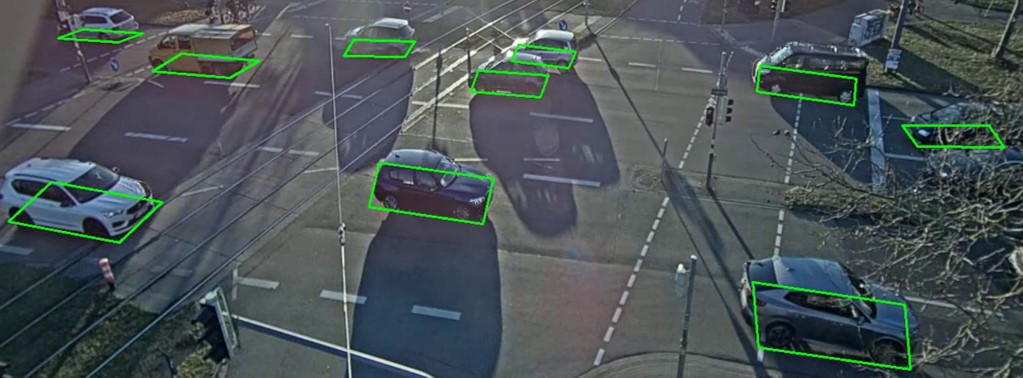}
\end{subfigure}

\vspace{0.05cm}

\begin{subfigure}[t]{1\linewidth}
  \includegraphics[width=1.0\textwidth]{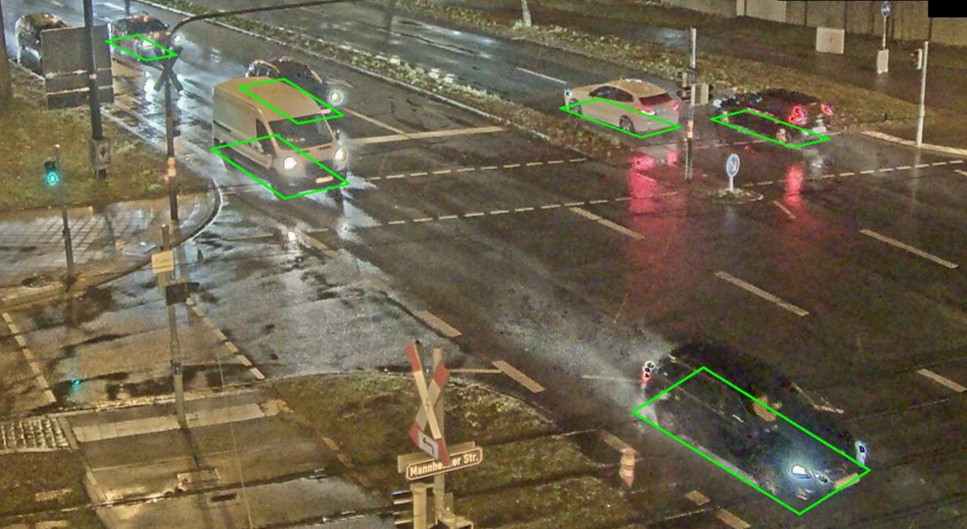}
\end{subfigure}

\caption{2.5D object detector predictions of vehicles for: \\ (i) real-world deployment; (ii) a held-out camera viewpoint; (iii) previously unobserved rain and nighttime conditions.}
\label{initimage}
\end{figure}

Autonomous vehicles primarily rely on on-board sensors such as cameras and Lidar for environmental perception. However, these sensors are inherently limited by occlusions and narrow fields of view, posing challenges in dynamic urban environments where relevant objects may not be visible. This is particularly critical at traffic intersections, where autonomous driving systems must explicitly handle non-observable regions to ensure safe driving maneuvers~\cite{schorner2019predictive, zipfl2025digit4taf}.
Unlike vehicle-mounted sensor systems, roadside infrastructure-based methods leverage elevated vantage points to capture wider fields of view with reduced occlusions and can transmit information to vehicles via V2X communication~\cite{zofka2022, zimmer2024tumtraf}. However, these elevated viewpoints also introduce new challenges. Although 3D object detection models perform well with on-board vehicle sensors, they often fail to generalize to elevated viewing angles. Furthermore, to reduce costs, intelligent infrastructure typically employs monocular cameras rather than Lidar, limiting the range of applicable 3D object detection methods. While autonomous vehicle perception has abundant datasets~\cite{Geiger2012CVPR, nuscenes, sun2020scalability}, infrastructure-based perception datasets are limited, necessitating efficient data collection and training strategies.
Current state-of-the-art methods exhibit several limitations, such as assuming uniform vehicle dimensions, enforcing alignment with lane directions and past trajectories, or relying on planar road surfaces~\cite{fleck2019towards, clausse2019large, kornfeld2024kalman, rezaei20233d, carrillo2021urbannet, zhu2021monocular}. To address these challenges, we present a real-time–capable 2.5D object detection framework. It can be trained efficiently with minimal real-world data, generalizes across diverse camera angles and weather conditions, and outperforms previous approaches. 

\section{Related Work}

\begin{figure}[t]
\centering

\begin{subfigure}[t]{1\linewidth}
  \includegraphics[width=1.0\textwidth]{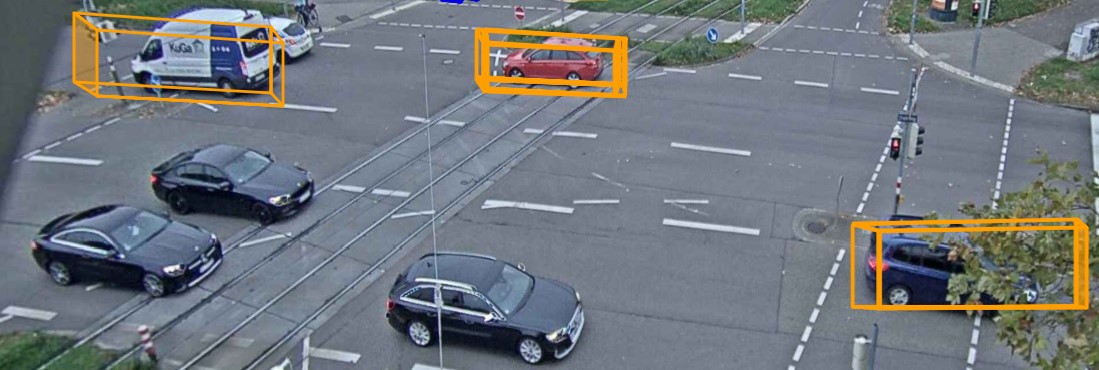}
\end{subfigure}
\caption{Monocular object detector FCOS3D~\cite{wang2021fcos3d} fails to generalize for elevated camera viewpoints. Vehicles are not detected, and predicted sizes and orientations are inaccurate.}
\label{fcos}
\end{figure}

\begin{figure*}[!h]
\centering
\begin{subfigure}[!t]{1\textwidth}
  \includegraphics[width=1.0\textwidth]{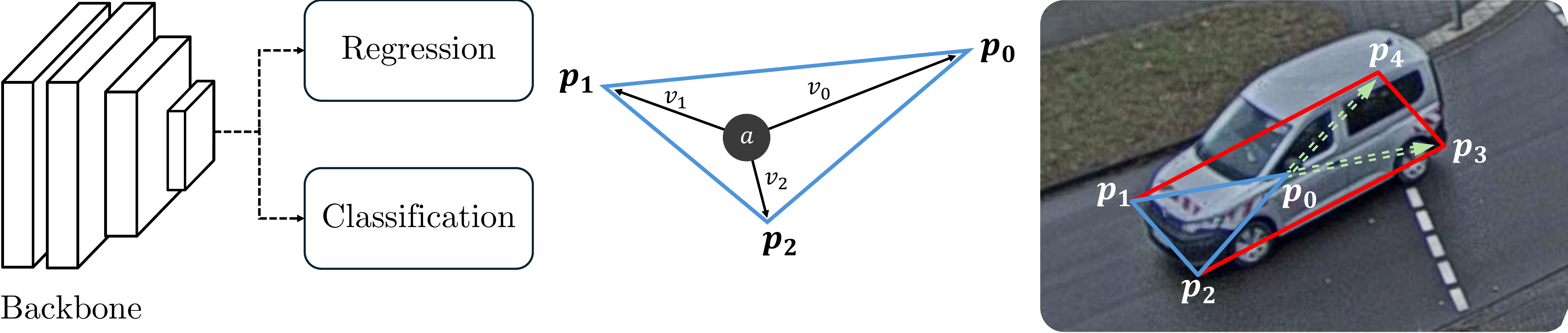}
\end{subfigure}
\caption{Regression for ground plane parallelogram prediction: The predicted vectors $v_1, v_2$ and $v_3$ are summed with anchor point $a$ to create triangle $p_0$, $p_1$, $p_2$. The parallelogram is obtained by reflecting $p_1$ and $p_2$ across $p_0$. }
\label{2.5Dframework}
\end{figure*}

\subsection{Camera-based Object Detection}
Camera-based object detection is typically categorized into 2D and 3D detection, both aiming to distinguish relevant objects from the background.
Standard 2D object detectors detect objects by predicting their $(x,y)$ position, width, and height in pixel coordinates to generate axis-aligned boxes, which are then classified. Most baseline architectures are trained and evaluated on the MS COCO dataset~\cite{lin2014microsoft} with large architectures, oftentimes based on Detection Transformer (DETR)~\cite{carion2020end}, achieving benchmark performance~\cite{zong2023detrs, zhang2022dino, chen2023group}. A separate line of research prioritizes inference speed to enable real-time object detection without significantly sacrificing accuracy. Here, YOLO-based models~\cite{redmon2016you,redmon2018yolov3,wang2023yolov7,jocher2023yolov8,wang2025yolov9,wang2024yolov10,yolo11_ultralytics,tian2025yolov12} and RT-DETR methods~\cite{zhao2024detrs,lv2024rt,wang2025rt} are among the most commonly employed approaches.

3D camera-based object detection detects objects in real-world 3D coordinates. These methods typically predict a rectangular cuboid, defined by the object's $(x,y,z)$ position, width, height, and length. The object's orientation, expressed in Euler angles, is also predicted. 
Because most 3D object detection datasets in the traffic domain require additional Lidar data for ground truth annotations, they are typically constructed using data collected from vehicles such as KITTI~\cite{Geiger2012CVPR}, nuScenes~\cite{nuscenes}, or Waymo~\cite{sun2020scalability}. In this domain, the majority of detected objects (vehicles and pedestrians) are situated on a mostly planar road surface, allowing for the common simplification of assuming zero pitch and roll angles.
Monocular 3D object detection is an inherently ill-posed problem, as 2D images introduce scale ambiguities that must be approximated by inferring depth and learning geometric priors of frequently encountered objects~\cite{mao20233d}.

One monocular approach, M3D-RPN~\cite{brazil2019m3d}, combines 2D and 3D region proposal networks using 3D anchors to create strong priors correlating 2D scale with 3D depth. For orientation, only the yaw angle is predicted, and it is further refined by reprojecting the 3D cuboid into the image plane and rotating objects to maximize overlap with 2D detections. Another approach, FCOS3D~\cite{wang2021fcos3d}, extends the anchor-free FCOS~\cite{tian2019fcos} 2D object detection framework to monocular 3D object detection by projecting the 3D object centers onto the image plane and decoupling 2D attributes from depth and orientation estimates.


\subsection{Object Detectors for Roadside Infrastructure}
Intelligent roadside infrastructure, whether aimed at enhancing or validating autonomous systems or managing traffic flow, requires precise and reliable object detection capabilities.
Off-the-shelf 2D object detectors, e.g., YOLO-architectures, exhibit strong generalization capabilities to steep camera viewpoints, largely due to training on large and diverse datasets. Consequently, many existing approaches leverage 2D object detectors in combination with projection techniques to approximate 3D bounding boxes. However, off-the-shelf monocular 3D object detectors do not generalize well due to significant domain shifts, exemplified in Fig.~\ref{fcos}.

\cite{fleck2019towards}, \cite{clausse2019large}, \cite{kornfeld2024kalman}, and  \cite{rezaei20233d}  rely on 2D object detections projected from image frame into world coordinates. They assume fixed object sizes and estimate orientation from past trajectories. These approaches tackle the ill-posed nature of monocular 3D detection by relying on geometric approximations and precise camera calibration.
For example, Rezaei et al.~\cite{rezaei20233d} employ approximately 70 images per camera, along with top-view satellite imagery, to perform inverse perspective mapping~\cite{mallot1991inverse} and project 2D detections into 3D. They assign a fixed width and length to each object class, which is used for depth estimation of 2D object detections. To estimate the object's heading (yaw angle), they track past detections and approximate the orientation based on the object's motion trajectory.  Since trajectory-based orientation estimation fails for static objects, they also assume parallel heading relative to known road lanes. They evaluate their method on 2D datasets~\cite{luo2018mio, wen2020ua, guerrero2013iwinac} and the non-public Leeds~\cite{rezaei20233d} dataset.
\newline UrbanNet~\cite{carrillo2021urbannet} similarly uses 2D object detection but additionally incorporates exact 3D road geometry to aid in the projection, assuming vehicles are positioned at the center of the road. Compared to the previous methods, this allows projections for elevated and sloped roads. This approach requires precise knowledge of the 3D road geometry, which is difficult to obtain in practice, resulting in an evaluation limited to synthetically generated data.
\newline YOLOv7-3D~\cite{ye2023yolov7} foregoes the projection approach and directly trains  a 3D object detector on the Rope3D~\cite{ye2022rope3d} dataset. At the time of writing, the Rope3D dataset is only available in China and, therefore, inaccessible for this work. Their evaluation reports low Average Precision~(AP) scores. 
\newline Zhu et al.~\cite{zhu2021monocular}, similarly to Rezaei et al.~\cite{rezaei20233d}, annotate keypoints in satellite and roadside camera images and compute a homography mapping images of roadside cameras into a bird's-eye view (BEV) assuming planar roads. 2D object detection is subsequently applied to these BEV images, estimating rotated bounding boxes, which correspond to the yaw-angle orientation. However, transforming roadside images into BEV space creates large distortions.

In contrast to the methods described above, our approach addresses the ill-posed depth estimation problem without relying on fixed vehicle sizes or the assumption of planar road geometry. Instead, we directly learn depth and orientation through a novel 2.5D object detection technique.
\newpage

\section{Method}
Object detection for intelligent roadside infrastructure cameras is primarily utilized for traffic flow management and perception systems of autonomous vehicles~\cite{amirgholy2020optimal, fleck2019towards, clausse2019large}.
We argue that for these applications, object height adds little value, whereas precise localization on the road surface and orientation are critical. Methods that infer orientation from explicit road geometries or past trajectories~\cite{carrillo2021urbannet, rezaei20233d, fleck2019towards} perform well in typical scenarios but often fail in complex maneuvers such as U-turns, lane changes, or with stationary vehicles. Therefore, to improve efficiency and reduce complexity, we reformulate the 3D object detection task as a 2.5D task by omitting object height and focusing solely on detecting each participant's ground-plane footprint. For vehicles and bicycles, this footprint appears as a parallelogram in the image under an approximately affine projection, an assumption that holds for many roadside cameras.  In this work, we propose a model to detect these parallelograms directly in image space. This approach eliminates the need for camera calibration and trajectory tracking, simplifies the detection task compared to full 3D methods, and preserves the essential spatial information for downstream applications.
While the method could, in principle, be extended to pedestrians, their ground-plane projections are not well captured by a parallelogram approximation. Consequently, this work focuses exclusively on vehicle detection and does not address pedestrians.
\subsection{Dataset Creation}
\begin{figure}[t]
\centering

\begin{subfigure}[t]{1\linewidth}
  \includegraphics[width=1.0\textwidth]{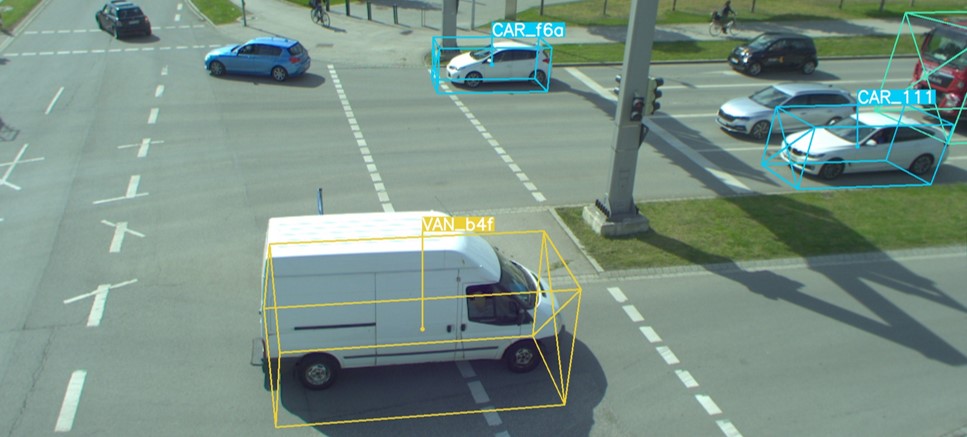}
\end{subfigure}
\caption{TUMTraf~\cite{zimmer2023tumtrafintersection} annotations are often misaligned or missing.}
\label{TumTraf}
\end{figure}

For training the proposed 2.5D object detector, datasets containing annotated parallelograms are required. 
However, to the best of our knowledge, no publicly available dataset is dedicated to ground-plane estimation from roadside cameras. Popular datasets like MIO-TCD~\cite{luo2018mio}, UA-DETRAC~\cite{wen2020ua}, and GRAM-RTM~\cite{guerrero2013iwinac} provide only 2D bounding box information and therefore lack the geometric information needed to recover ground planes. Conversely, 3D datasets are suitable in principle, as the required ground plane information can be calculated by projecting the four bottom vertices of each 3D bounding box into the image plane.
Since the Rope3D~\cite{ye2022rope3d} and Leeds~\cite{rezaei20233d} datasets are inaccessible, we examine the TUMTraf~\cite{zimmer2024tumtraf,zimmer2023tumtrafintersection} dataset for potential use. While camera viewpoints are suitable, we observe that a significant portion of annotations are inaccurate (cf. Figure~\ref{TumTraf}), limiting the dataset's direct applicability to training our model.

To obtain reliable data, we recorded a new dataset with the intelligent roadside infrastructure of the Test Area Autonomous Driving Baden-Württemberg (TAF-BW)~\cite{fleck2019towards,zipfl2020traffic,fleck2023semi}. We captured 3,633 images using seven roadside cameras. Sampling images at 5-minute intervals increased the diversity of traffic situations. Each image was then manually annotated. As the labeling software CVAT~\cite{cvat} lacks native support for parallelogram annotations, we followed the L-shape, triangle-based approach proposed by Fleck et al.~\cite{fleck2023semi}: By annotating three outermost vehicle corners, the fourth corner is inferred by reflecting one point across the diagonal, exploiting parallelogram geometry.
Since the TUMTraf images provide valuable viewpoint diversity, we selected 282 images from this dataset. Due to their poor annotation quality, we discarded the original object annotations and manually created new parallelogram labels according to the previously described procedure.
Even with the efficient triangle-based annotation approach, manual labeling is resource-intensive, so the size of the real-world dataset is small. Furthermore, the limited number of available roadside cameras restricts the diversity of viewpoints. To enrich the dataset variety, we employ the simulation engine CARLA~\cite{dosovitskiy2017carla}. We manually placed 818 unique camera positions spread across seven CARLA maps and utilized CARLA-provided ground truth 3D bounding boxes for all spawned vehicles. We obtained the 2.5D ground plane labels by selecting the four bottom vertices, discarding the top vertices. To mitigate bounding box lag, CARLA's synchronous mode was utilized. Table~\ref{tab:dataset_overview} summarizes the dataset composition used for training the proposed 2.5D object detector.

\begin{table}[h]
\centering
    \caption{Overview of the dataset composition, including total number of images, camera viewpoints, and labeled objects per dataset.}
    \label{tab:dataset_overview}

    \begin{threeparttable}[b]
    \centering
    \resizebox{1.0\linewidth}{!}{

        \begin{tabular}{r cc c}
            \toprule
            & \multicolumn{2}{c}{\textbf{Real-world}} & \textbf{Synthetic} \\
            \cmidrule(lr){2-3} \cmidrule(lr){4-4}
            & \textbf{TAF-BW} & \textbf{TUMTraf} & \textbf{CARLA} \\ 
            \midrule
            \textbf{Images} & 3,633 & 282 & 218,223 \\ 
            \textbf{Viewpoints} & 7 & 4 & 818 \\ 
            \textbf{Objects} & 20,678 & 1,893 & 1,534,474 \\ 
            \bottomrule
        \end{tabular}
    }
    \end{threeparttable}

\end{table}

\newpage
\subsection{2.5D Object Detector}
Given a monocular image from a single roadside camera, the object detector's task is to predict the ground plane of all vehicles. 2D/3D object detectors typically leverage their rectangular bounding box structure to regress the center, width, height, and, in the case of 3D, length and rotation. In a similar approach, a parallelogram could be parameterized by six values $(x,y), w, h, \theta, \alpha$. Here, $(x,y)$ denotes the center, $w$ and $h$ the side lengths, $\theta$ the rotation, and $\alpha$ the internal slant angle. Yet $\alpha$ alone is ambiguous, as a parallelogram contains two distinct internal angles, each occurring at a pair of opposite vertices. Therefore, we would also need to specify the vertex for which $\alpha$ is predicted.
To address this limitation, we propose a representation based on the geometric property that a parallelogram defined by vertices $(p_1, p_2, p_3, p_4)$ and center point $p_0$ can be reconstructed if $(p_0, p_1, p_2)$ are known, by mirroring $p_1$ and $p_2$ across $p_0$ to create $p_3$ and $p_4$. Thus, predicting only $p_0, p_1,$ and $p_2$ suffices to reconstruct the full parallelogram. Moreover, flipping the points $p_1$ and $p_2$ does not change the resulting parallelogram. Consequently, our network regresses a triangle composed of the center and the two front corners, rather than the entire parallelogram. Accordingly, the network outputs six values per vehicle detection, specifically the $(x,y)$ coordinates of the points $p_0$, $p_1$, and $p_2$.

\subsubsection{Neural Network Architecture}
We base our model on the 2D object detector YOLOv8~\cite{jocher2023yolov8} and adapt its regression head to enable triangle prediction. The classification head remains unchanged. Figure~\ref{2.5Dframework} illustrates our modified regression scheme for parallelogram prediction.
For regression, we adopt an anchor-box-free design: each pixel in the final feature maps serves as an anchor point, for which offsets are added to predict the triangle vertices. In the 2D anchor-free regression~\cite{tian2019fcos, jocher2023yolov8}, offsets to the anchor points are used to determine the width, height, and position of the rectangular bounding box. In the case of arbitrary triangles, we directly sum the predicted vector offsets to the anchor point. This allows for arbitrary triangle sizes and orientations, eliminating the requirement to predefine how many vertices should be positioned to the left or right of the anchor point. In the 2D approach, all anchor points located inside a ground truth bounding box are responsible for predicting the bounding box, creating a one-to-many label assignment. It assumes that every anchor point inside the ground-truth box is located within the object in the image frame. For our proposed triangle prediction, this assumption does not hold, as the parallelogram ground plane exceeds the size of the triangle. We therefore propose a tolerance $\eta$, such that anchor points outside the triangle region are still considered active and contribute to predicting the triangle. This is possible because summing the predicted vector offsets allows triangle prediction where the anchor point is not contained within the triangle. 

\subsubsection{Loss Function for Regression}
\begin{figure}[t]
\centering

\begin{subfigure}[t]{1\linewidth}
  \includegraphics[width=1.0\textwidth]{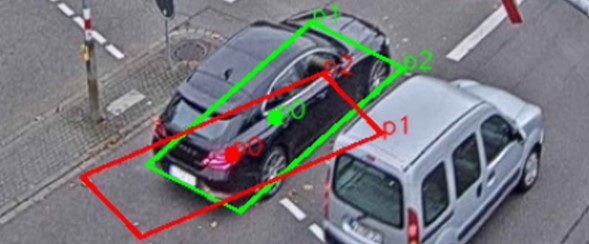}
\end{subfigure}
\caption{Example from early training: Prediction in red, ground truth in green. $p_1$ and $p_2$ are flipped between ground truth and prediction. In this case, the regression loss calculates the distance between $p_1$ and $\hat p_2$ and $p_2$ with $\hat p_1$.}
\label{Chamfer}
\end{figure}

Shifting from a rectangular bounding box to a triangle regression introduces new loss-design challenges. The Complete Intersection over Union (CIoU) loss function~\cite{zheng2020distance} used in most modern 2D object detection methods is not well-suited for the task of 2.5D object detection. While IoU can be computed quickly for axis-aligned 2D rectangles, calculating IoU for triangles requires the Shoelace formula and Sutherland-Hodgman clipping, which significantly decreases training speed. Further, IoU loss functions do not consider the order of predicted points. In our case, the distinction between the center point $p_0$ and the vertices $p_1$ and $p_2$ is crucial, as the center point is used to mirror the other points to create the final parallelogram. 
We therefore employ Mean Squared Error (MSE) to compare the predicted center point $p_0$ with the ground truth center point $\hat p_0$. 
 For the vertices $p_1$ and $p_2$, however, only the \textit{unordered} set of the points matters; interchanging them leaves the final parallelogram unchanged. MSE, by construction, compares points in a fixed order, so if the network predicts $p_1$ where the label lists $\hat p_2$ (and vice-versa), the loss is computed as two large squared distances, even though the resulting parallelogram is identical. Consequently, MSE is unsuitable for predicting $p_1$ and $p_2$, and a permutation invariant loss is required. Instead, we use the Chamfer Distance: 
$$
    \mathcal{L}_{C D}=\frac{1}{|P|} \sum_{p \in P} \min _{q \in Q}\|p-q\|_2^2+\frac{1}{|Q|} \sum_{q \in Q} \min _{p \in P}\|q-p\|_2^2,
$$
where $P$ represents the set of predicted triangle vertices $p_1$ and $p_2$ and $Q$ is the set of ground truth triangle vertices $\hat p_1$ and $\hat p_2$ . 
This allows the model to predict these two vertices in either clockwise or counter-clockwise order, reflecting the symmetry of the parallelogram. In our experiments, using the Chamfer Distance instead of MSE for $p_1$ and $p_2$ stabilizes early training, as qualitatively showcased in Fig.~\ref{Chamfer}. The final regression loss is the unweighted sum of MSE and the Chamfer Distance.

\subsection{Training of 2.5D Object Detector}
Our dataset contains real-world images (TAF-BW and TUMTraf) and synthetic images generated with CARLA. 
We evaluate four training strategies:
\begin{itemize}

    \item Train exclusively on real-world data.
    \item Train exclusively on synthetic data to quantify the synthetic-to-real domain gap.
    \item Pretrain on synthetic data and subsequently finetune on real-world data.
    \item Pretrain on synthetic data and subsequently finetune on a mix of 70 \% real-world and 30 \% synthetic data.
    
\end{itemize}

Because we alter only the regression head of the YOLOv8 baseline, we can retain the backbone and classification layers pretrained on MS COCO~\cite{lin2014microsoft} or Open Images v7 (OIV7)~\cite{OpenImages}. Both datasets already contain vehicle classes, so the transferred weights might provide a good weight initialization for 2.5D object detection and accelerate training. 
During pretraining with synthetic data, we use the SGD optimizer and train for 9 epochs.  
Finetuning runs for 120 epochs on real-world data or 70 epochs on the mixed set, starting at a learning rate of 0.001 that decays to 0.0001 with cosine scheduling. Following~\cite{ruis2024improving}, we heavily augment images to narrow the domain gap between synthetic and real-world images.

\subsection{Inference and Deployment}
While in training only three vertices are regressed, in inference we obtain the full parallelogram by reflecting the two predicted non-center vertices across the predicted center point. The proposed regression head and loss lead to many predictions of overlapping parallelograms because, during training, the one-to-many label assignment maps multiple predictions to the same ground truth. Typically, this is resolved by applying non-maximum suppression (NMS), discarding low-confidence detections that have a high IoU with high-confidence detections. As detailed earlier, calculating IoU for triangles or parallelograms is computationally intensive, so we replace it with an inexpensive approximation during NMS. For each predicted parallelogram, we enclose it in an axis-aligned rectangle derived from its vertices and calculate the IoU for these rectangles, significantly reducing computational complexity. In testing, the average discrepancy between the exact IoU and the approximated IoU was 0.05, which, for reducing total inference time by a factor of three, is an acceptable approximation error.

Deploying the FP16-TensorRT model on TAF-BW hardware (NVIDIA T4 GPU) yields 39.2~ms inference time per image, supporting real-time object detection.
We reduce the class confidence threshold to 0.1, distinguishing between background and object detections, as false positives are rare and recall increases. 
\newpage
\section{Evaluations}

\begin{table*}[t]
    \caption{Detection performance on three evaluation test sets.
             Columns 1–3 list weight initialization, pretraining, and finetuning datasets. 
             The remaining columns report Precision (P), Recall (R),
             mean Average Precision at 50\,\% IoU (mAP@50), Average IoU (AIoU)
             and mean Absolute Orientation Error (mAOE) for each test set.}
    \label{tab:all_test_results}
    \centering
    \setlength{\tabcolsep}{5pt}
    \resizebox{\textwidth}{!}{%
    \begin{tabular}{
        c c c                              
        | *{5}{c}                          
        | *{5}{c}                          
        | *{5}{c}                         
    }
        \toprule
        \makecell[c]{\textbf{Weight}\\\textbf{Init.}} &
        \textbf{Pretrain} &
        \textbf{Finetune} &
        \multicolumn{5}{c}{\textbf{Default Test}} &
        \multicolumn{5}{c}{\textbf{Camera Hold-Out Test}} &
        \multicolumn{5}{c}{\textbf{Night-Rain Test}}\\
        \cmidrule(lr){4-8}\cmidrule(lr){9-13}\cmidrule(lr){14-18}
            &   &   &
            P\up & R\up & mAP@50\up & AIoU\up & mAOE\down &
            P\up & R\up & mAP@50\up & AIoU\up & mAOE\down &
            P\up & R\up & mAP@50\up & AIoU\up & mAOE\down \\
        \midrule
        OIV7 & --    & Real-World           & 0.97 & 0.96 & 0.97 & 0.83 & 3.9 & 0.76 & 0.88 & 0.82 & 0.71 & 7.9 & 0.68 & 0.55 & 0.62 & 0.68 & 4.2 \\
        None & --    & Real-World            & 0.94 & 0.86 & 0.90 & 0.78 & 6.1 & 0.43 & 0.35 & 0.39 & 0.64 & 18.9 & 0.07 & 0.01 & 0.03 & 0.64 & 5.0 \\
        COCO & --    & Real-World            & 0.97 & 0.96 & 0.97 & 0.83 & 4.1 & 0.59 & 0.86 & 0.73 & 0.72 & 6.9 & 0.79 & 0.62 & 0.71 & 0.67 & 12.3 \\
        OIV7 & Synth. & --              & 0.90 & 0.92 & 0.91 & 0.73 & 7.9 & 0.67 & 0.85 & 0.76 & 0.66 & 3.9 & 0.87 & 0.65 & 0.76 & 0.64 & 4.3 \\
        None & Synth. & --              & 0.93 & 0.85 & 0.89 & 0.72 & 7.8 & 0.77 & 0.77 & 0.77 & 0.65 & 5.2 & 0.91 & 0.30 & 0.60 & 0.66 & 5.4 \\
        COCO & Synth. & --              & 0.91 & 0.92 & 0.92 & 0.73 & 6.8 & 0.63 & 0.84 & 0.74 & 0.66 & 3.2 & 0.89 & 0.71 & 0.80 & 0.63 & 4.8 \\
        OIV7 & Synth. & Real-World            & 0.97 & 0.96 & \textbf{0.97} & \textbf{0.85} & \textbf{3.1} & 0.79 & 0.85 & 0.82 & \textbf{0.73} & 3.5 & 0.99 & 0.74 & \textbf{0.86} & 0.73 & 3.8 \\
        None & Synth. & Real-World            & 0.98 & 0.96 & \textbf{0.97} & 0.82 & 3.5 & 0.81 & 0.83 & 0.82 & 0.71 & 3.4 & 0.86 & 0.41 & 0.64 & 0.72 & 4.4 \\
        COCO & Synth. & Real-World            & 0.97 & 0.96 & \textbf{0.97} & 0.83 & 3.2 & 0.79 & 0.86 & \textbf{0.83} & 0.72 & \textbf{2.3} & 0.98 & 0.65 & 0.81 & \textbf{0.74} & 3.8 \\
        OIV7 & Synth. & Mix  & 0.98 & 0.96 & \textbf{0.97} & 0.84 & \textbf{3.1} & 0.82 & 0.85 & \textbf{0.83} & 0.72 & 3.2 & 0.97 & 0.66 & 0.82 & 0.72 & \textbf{3.6} \\
        None & Synth. & Mix  & 0.98 & 0.94 & 0.96 & 0.82 & 4.1 & 0.84 & 0.83 & \textbf{0.83} & 0.71 & 2.9 & 0.93 & 0.37 & 0.65 & 0.71 & 3.9 \\
        COCO & Synth. & Mix  & 0.97 & 0.96 & \textbf{0.97} & 0.83 & 3.2 & 0.75 & 0.85 & 0.80 & 0.71 & 2.8 & 0.96 & 0.66 & 0.81 & 0.73 & 4.1 \\
        \bottomrule
    \end{tabular}}%
\end{table*}

Intelligent roadside infrastructure demands highly accurate detections to support downstream perception tasks such as validation of autonomous vehicles. We therefore evaluate the proposed method with respect to three objectives:
\begin{enumerate}
    \item Identifying vehicles in the scene
    \item Estimating vehicle positions accurately
    \item Determining vehicle orientation precisely
\end{enumerate}

\subsection{Evaluation Metrics}
To quantify the first objective, we report mAP@50, and for a more detailed error analysis, precision and recall. For the second objective, the Average Intersection-over-Union (AIoU) metric quantifies the degree of overlap between predicted and ground truth parallelograms.

\[
\mathrm{AIoU} \;=\; \frac{1}{\lvert\mathcal{M}\rvert} \sum_{(p,g)\in\mathcal{M}} \operatorname{IoU}(p,g),
\]

where $\mathcal{M}$ denotes the set of prediction–ground truth parallelogram pairs whose intersection-over-union exceeds 0.5, and $\lvert\mathcal{M}\rvert$ its cardinality. 
In contrast to mAP@50, the AIoU measures the exact overlap between prediction and ground truth for correctly classified predictions and is solely used to evaluate regression performance. 

\begin{figure}[t]
\centering
\begin{subfigure}[t]{1\linewidth}
  \includegraphics[width=1.0\textwidth]{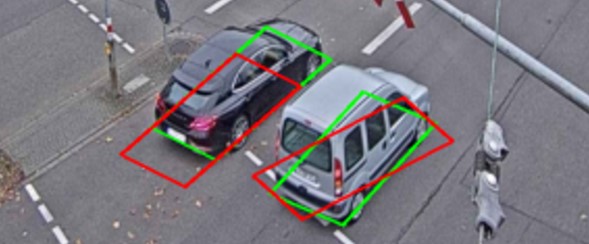}
\end{subfigure}
\caption{Predictions in red, annotations in green. The right vehicle prediction has only a small IoU error to its label. mAOE can be used to still extract the high orientation error. In contrast, the left vehicle prediction has a higher IoU error with small mAOE error.}
\label{orientation-metric}
\end{figure}

IoU-based metrics are inadequate for orientation evaluations, as large orientation errors can still lead to high IoU values (cf. Fig.~\ref{orientation-metric}). Inspired by 3D object detection metrics, we utilize the mean absolute orientation error (mAOE).
For each predicted parallelogram, we take the midpoint $x_1,y_1$ of its front edge and the midpoint $x_2,y_2$ of its rear edge. The same procedure for the ground-truth box yields $\hat{x}_1,\hat{y}_1$ and $\hat{x}_2,\hat{y}_2$. Using these four points, the absolute orientation angle error is calculated by 
$$
AOE=\left|\arctan \left(\frac{y_2-y_1}{x_2-x_1}\right)-\arctan \left(\frac{\hat y_2-\hat y_1}{\hat x_2 - \hat x_1}\right)\right|.
$$
The mAOE is calculated by averaging the AOE across all prediction-ground truth pairs.

\subsection{Evaluation Test Datasets}
To assess our trained models, we record and annotate three dedicated test sets using the roadside cameras of TAF-BW: 
\begin{enumerate}[label=(\roman*)]
    \item  \textbf{Default Test}: Images are recorded by the same seven cameras used during training but on different days. This set represents the typical deployment conditions expected within TAF-BW.
    \item \textbf{Camera Hold-Out Test}: Although TAF-BW provides eight cameras, only seven are included in the training set. The eighth camera, kept entirely unseen during training, serves to evaluate generalization to a new viewpoint.
    \item \textbf{Night-Rain Test}: The real-world training images comprise of only daylight scenes in overcast or sunny conditions. To evaluate robustness to different lighting and weather conditions, this test set consists of images recorded during nighttime and rain. 

\end{enumerate}

Table~\ref{tab:all_test_results} summarizes the results obtained with the different training strategies, while Figure~\ref{initimage} shows exemplary predictions for images of each test set. 

On the Default Test, models trained using the same camera viewpoints achieve higher mAP@50, AIoU, and lower (better) mAOE results, demonstrating effectiveness in familiar conditions. Models trained purely on synthetic data perform worse. Examining the mAOE, which represents the average angle difference to ground truth, a similar trend emerges: models trained on identical camera viewpoints yield higher orientation accuracies. The uninitialized model trained only on real data lags behind COCO- or OIV7-initialized versions, underscoring the value of weight initialization in low-data regimes. This difference is less pronounced for models trained with the large-scale synthetic pretraining dataset (2 million images).
The best models deliver high classification, localization, and orientation scores and are well-suited for further downstream tasks. 

For the Camera Hold-Out Test, a significant drop in the mAOE metric is observed for models not trained on synthetic data. This highlights the worse generalization capability of models trained solely on limited real-world data for estimating orientation on unseen road layouts.
Models trained solely on real data still benefit markedly from COCO or OIV7 initialization, outperforming the model trained from scratch.
Compared to the Default Test, mAP@50 and AIoU degrade in this test set for all models. Nonetheless, the leading models achieve high detection accuracy and maintain strong IoU and orientation consistency. This demonstrates reliable generalization to unseen viewpoints.

In the Night-Rain Test, models without synthetic pretraining data experience a sharp drop in performance. In particular, the uninitialized, non-pretrained model is unable to generalize to these new lighting conditions. Moreover, for all other training regimes, initializing weights from scratch reduces mAP@50 values, indicating significant challenges in classification quality under these unseen conditions.
Initializing weights with COCO or OIV7 and training on synthetic and real-world data achieves the highest results. Similarly to the Camera Hold-Out Test, all metrics decline relative to the Default Test. Nevertheless, top models still deliver high performance and generalize well. 

In summary, in most cases, initializing the backbone and classifier with Open Images V7 or MS COCO generally outperforms training from scratch. Especially in the tests designed to evaluate generalizability, initialized models achieve higher mAP@50 values, emphasizing the need for a robust initial feature representation to facilitate adaptation to new conditions. Initializing with either COCO or OIV7 is inconclusive, as no clear improvement of one initialization over the other was consistently observed.
Pretraining using the synthetically generated CARLA training dataset also significantly improves performance in the test sets requiring higher generalization capabilities. We observe no consistent advantage in finetuning on only real-world data compared to using a mixed real-world and synthetic dataset. Here, different tests favor different finetuning data. Overall, finetuning on real data alone slightly edges out the mixed strategy. 
For deployment, we use the model initialized with OIV7 weights, pretrained on synthetic data, and finetuned on real-world data only. This configuration achieves consistently strong performance across all test sets and is therefore adopted. To differentiate it from the previous TAF models, we call it TAF-v2.5. All detections in Fig.~\ref{initimage} were created by this model.

\subsection{Evaluation Against 2D Object Detection with Projection}
We hypothesize that 2.5D detection offers superior localisation and orientation accuracy compared with 2D object detection approaches that project their rectangular bounding boxes into world coordinates. We therefore evaluate TAF-v2.5 against a similar model to the one detailed in~\cite{fleck2019towards}, calling it TAF-v1. Because TAF-v1 employs a tracker for object orientation, it must be evaluated on video streams rather than on individual frames. We therefore annotated a video stream at 15~Hz to enable a fair comparison. 
Note that we evaluate only vehicles, although TAF-v1 can also detect pedestrians. We also compare only TAF-v1's ground-plane projection. This remains fair because TAF-v1 assumes a fixed vehicle height anyway; extending TAF-v2.5 outputs with the same height would yield identical 3D bounding box comparisons. 

\begin{figure}[t]
\centering
\begin{subfigure}[t]{0.49\linewidth} 
    \includegraphics[width=1.0\textwidth]{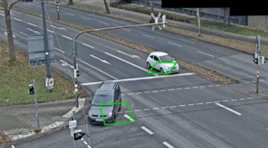}
    \caption{Predictions of TAF-v1.}
\end{subfigure}
\hfill 
\begin{subfigure}[t]{0.49\linewidth} 
  \includegraphics[width=1.0\textwidth]{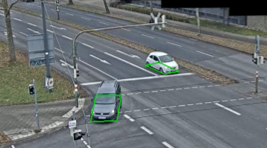}
  \caption{Predictions of TAF-v2.5.}
\end{subfigure}
\caption{Qualitative comparison of TAF-v1 with TAF-v2.5. In this scenario, the white vehicle is stationary and the gray vehicle is turning right. For TAF-v1, the predicted orientation for both vehicles is inaccurate, as past trajectories cannot be relied on for orientation assumptions.}
\label{tafvs25}
\end{figure}

Table~\ref{tab:taf_comparison} compares the performance of the models on this video stream.
Both models show high mAP@50 scores, signifying good classification and approximate localization results. Significant differences are observed in AIoU and mAOE. While the 2D projection approach only roughly approximates the localisation and orientation of detected vehicles, the 2.5D approach demonstrates significantly higher performance. Figure~\ref{tafvs25} showcases an example in which these differences are displayed.

\begin{table}[t]
    \centering
    \caption{Performance comparison between the TAF-v2.5 model and the TAF-v1 model similar to~\cite{fleck2019towards}.}
    \label{tab:taf_comparison}           
    \begin{threeparttable}[b]
    \resizebox{1.0\linewidth}{!}{
    \centering

        \begin{tabular}{l c c c c c}
            \toprule
            & P\up & R\up & mAP@50\up & AIoU\up & mAOE\down\\
            \midrule
            TAF-v2.5     & 0.99 & 0.99 & 0.99 & 0.93 & 1.2  \\
            TAF-v1 & 0.85 & 0.95 & 0.90 & 0.39 & 34.2 \\
            \bottomrule
        \end{tabular}
        }
    \end{threeparttable}
\end{table}

\section{Conclusion}
In this work, we presented a 2.5D object detection framework adapted to the unique geometry and operating constraints of roadside infrastructure cameras.
By reformulating detections as direct predictions of ground-plane parallelograms and implementing this via a triangle-based regression head, we eliminate the need for calibration, road-layout priors, and trajectory tracking that limited earlier approximative projection-based approaches. To compensate for the scarcity and viewpoint bias of existing roadside datasets, we introduced a hybrid training strategy that combines synthetically generated and real-world images. Evaluation on three distinct test datasets shows that our TAF-v2.5 model yields strong generalisation capabilities to unseen viewpoints and weather conditions, while maintaining real-time performance.  Compared to the previous TAF-v1 2D-projection baseline, our approach achieves higher localization and orientation accuracy, confirming the benefits of direct ground-plane prediction. Model weights and code are publicly available to encourage future research in intelligent roadside infrastructure perception.

\vspace{0.70cm}
\section*{Acknowledgment}

This work was supported by funding from the Topic Engineering Secure Systems of the Helmholtz
Association (HGF) and by KASTEL Security Research Labs (46.23.03). 
It was also supported under the 'AI in Mobility' service contract, on behalf of the Ministry of Transport Baden-Württemberg.


{\small
\bibliographystyle{IEEEtran}
\bibliography{references}

\begin{thebibliography}{10}
\providecommand{\url}[1]{#1}
\csname url@samestyle\endcsname
\providecommand{\newblock}{\relax}
\providecommand{\bibinfo}[2]{#2}
\providecommand{\BIBentrySTDinterwordspacing}{\spaceskip=0pt\relax}
\providecommand{\BIBentryALTinterwordstretchfactor}{4}
\providecommand{\BIBentryALTinterwordspacing}{\spaceskip=\fontdimen2\font plus
\BIBentryALTinterwordstretchfactor\fontdimen3\font minus \fontdimen4\font\relax}
\providecommand{\BIBforeignlanguage}[2]{{%
\expandafter\ifx\csname l@#1\endcsname\relax
\typeout{** WARNING: IEEEtran.bst: No hyphenation pattern has been}%
\typeout{** loaded for the language `#1'. Using the pattern for}%
\typeout{** the default language instead.}%
\else
\language=\csname l@#1\endcsname
\fi
#2}}
\providecommand{\BIBdecl}{\relax}
\BIBdecl

\bibitem{schorner2019predictive}
P.~Sch{\"o}rner, L.~T{\"o}ttel, J.~Doll, and J.~M. Z{\"o}llner, ``{Predictive Trajectory Planning in Situations with Hidden Road Users Using Partially Observable Markov Decision Processes},'' in \emph{Intelligent Vehicles Symposium (IV)}.\hskip 1em plus 0.5em minus 0.4em\relax IEEE, 2019.

\bibitem{zipfl2025digit4taf}
M.~Zipfl, P.~Zwick, P.~Schulz, M.~R. Zofka, A.~Schotschneider, H.~Gremmelmaier, N.~Polley, F.~M{\"u}tsch, K.~Simon, F.~Gottselig \emph{et~al.}, ``{DigiT4TAF--Bridging Physical and Digital Worlds for Future Transportation Systems},'' in \emph{International Automated Vehicle Validation Conference (IAVVC)}.\hskip 1em plus 0.5em minus 0.4em\relax IEEE, 2025.

\bibitem{zofka2022}
M.~R. Zofka, T.~Fleck, and J.~M. Zöllner, ``{A Unified Description of Proving Grounds and Test Areas for Automated and Connected Vehicles},'' in \emph{Intelligent Vehicles Symposium (IV)}.\hskip 1em plus 0.5em minus 0.4em\relax IEEE, 2022.

\bibitem{zimmer2024tumtraf}
W.~Zimmer, G.~A. Wardana, S.~Sritharan, X.~Zhou, R.~Song, and A.~C. Knoll, ``{TUMTraf V2X Cooperative Perception Dataset},'' in \emph{Conference on Computer Vision and Pattern Recognition (CVPR)}, 2024.

\bibitem{Geiger2012CVPR}
A.~Geiger, P.~Lenz, and R.~Urtasun, ``{Are we ready for Autonomous Driving? The KITTI Vision Benchmark Suite},'' in \emph{Conference on Computer Vision and Pattern Recognition (CVPR)}, 2012.

\bibitem{nuscenes}
H.~Caesar, V.~Bankiti, A.~H. Lang, S.~Vora, V.~E. Liong, Q.~Xu, A.~Krishnan, Y.~Pan, G.~Baldan, and O.~Beijbom, ``{nuScenes: A multimodal dataset for autonomous driving},'' in \emph{Conference on Computer Vision and Pattern Recognition (CVPR)}, 2020.

\bibitem{sun2020scalability}
P.~Sun, H.~Kretzschmar, X.~Dotiwalla, A.~Chouard, V.~Patnaik, P.~Tsui, J.~Guo, Y.~Zhou, Y.~Chai, B.~Caine \emph{et~al.}, ``{Scalability in Perception for Autonomous Driving: Waymo Open Dataset},'' in \emph{Conference on Computer Vision and Pattern Recognition (CVPR)}.\hskip 1em plus 0.5em minus 0.4em\relax IEEE, 2020.

\bibitem{fleck2019towards}
T.~Fleck, K.~Daaboul, M.~Weber, P.~Sch{\"o}rner, M.~Wehmer, J.~Doll, S.~Orf, N.~Su{\ss}mann, C.~Hubschneider, M.~R. Zofka \emph{et~al.}, ``{Towards Large Scale Urban Traffic Reference Data: Smart Infrastructure in the Test Area Autonomous Driving Baden-W{\"u}rttemberg},'' in \emph{International Conference on Intelligent Autonomous Systems}.\hskip 1em plus 0.5em minus 0.4em\relax Springer, 2019.

\bibitem{clausse2019large}
A.~Clausse, S.~Benslimane, and A.~de~La~Fortelle, ``{Large-scale extraction of accurate vehicle trajectories for driving behavior learning},'' in \emph{2019 IEEE Intelligent Vehicles Symposium (IV)}.\hskip 1em plus 0.5em minus 0.4em\relax IEEE, 2019.

\bibitem{kornfeld2024kalman}
N.~Kornfeld, A.~Leich, and M.~Roth, ``Kalman filtering aspects in camera and deep learning based tracking for traffic monitoring,'' in \emph{International Conference on Information Fusion (FUSION)}.\hskip 1em plus 0.5em minus 0.4em\relax IEEE, 2024.

\bibitem{rezaei20233d}
M.~Rezaei, M.~Azarmi, and F.~M.~P. Mir, ``{3D-Net: Monocular 3D object recognition for traffic monitoring},'' \emph{Expert Systems with Applications}, vol. 227, 2023.

\bibitem{carrillo2021urbannet}
J.~Carrillo and S.~Waslander, ``{UrbanNet: Leveraging Urban Maps for Long Range 3D Object Detection},'' in \emph{International Conference on Intelligent Transportation Systems (ITSC)}, 2021.

\bibitem{zhu2021monocular}
M.~Zhu, S.~Zhang, Y.~Zhong, P.~Lu, H.~Peng, and J.~Lenneman, ``{Monocular 3D Vehicle Detection Using Uncalibrated Traffic Cameras through Homography},'' in \emph{International Conference on Intelligent Robots and Systems (IROS)}, 2021.

\bibitem{wang2021fcos3d}
T.~Wang, X.~Zhu, J.~Pang, and D.~Lin, ``{FCOS3D: Fully Convolutional One-Stage Monocular 3D Object Detection},'' in \emph{International Conference on Computer Vision (ICCV)}, 2021.

\bibitem{lin2014microsoft}
{T.-Y. Lin, M. Maire, S. Belongie et al.}, ``Microsoft {COCO}: Common objects in context,'' in \emph{European Conference on Computer Vision (ECCV)}.\hskip 1em plus 0.5em minus 0.4em\relax Springer, 2014.

\bibitem{carion2020end}
N.~Carion, F.~Massa, G.~Synnaeve, N.~Usunier, A.~Kirillov, and S.~Zagoruyko, ``{End-to-End Object Detection with Transformers},'' in \emph{European Conference on Computer Vision (ECCV)}.\hskip 1em plus 0.5em minus 0.4em\relax Springer, 2020.

\bibitem{zong2023detrs}
Z.~Zong, G.~Song, and Y.~Liu, ``{DETRs with Collaborative Hybrid Assignments Training},'' in \emph{International Conference on Computer Vision (ICCV)}, 2023.

\bibitem{zhang2022dino}
H.~Zhang, F.~Li, S.~Liu, L.~Zhang, H.~Su, J.~Zhu, L.~M. Ni, and H.-Y. Shum, ``{DINO: DETR with Improved DeNoising Anchor Boxes for End-to-End Object Detection},'' in \emph{International Conference on Learning Representations (ICLR)}, 2023.

\bibitem{chen2023group}
Q.~Chen, X.~Chen, J.~Wang, S.~Zhang, K.~Yao, H.~Feng, J.~Han, E.~Ding, G.~Zeng, and J.~Wang, ``{Group DETR: Fast DETR Training with Group-Wise One-to-Many Assignment},'' in \emph{International Conference on Computer Vision (ICCV)}, 2023.

\bibitem{redmon2016you}
J.~Redmon, S.~Divvala, R.~Girshick, and A.~Farhadi, ``{You Only Look Once: Unified, Real-Time Object Detection},'' in \emph{Conference on Computer Vision and Pattern Recognition (CVPR)}, 2016.

\bibitem{redmon2018yolov3}
{J. Redmon and A. Farhadi}, ``{YOLOv3: An Incremental Improvement},'' in \emph{arXiv preprint arXiv:1804.02767}, 2018.

\bibitem{wang2023yolov7}
C.~Wang, A.~Bochkovskiy, and H.~M. Liao, ``{YOLOv7: Trainable Bag-of-Freebies Sets New State-of-the-Art for Real-Time Object Detectors},'' in \emph{Conference on Computer Vision and Pattern Recognition (CVPR)}, 2023.

\bibitem{jocher2023yolov8}
\BIBentryALTinterwordspacing
G.~Jocher, A.~Chaurasia, and J.~Qiu, ``{YOLO by Ultralytics (Version 8.0.0)},'' 2023. [Online]. Available: \url{https://github.com/ultralytics/ultralytics}
\BIBentrySTDinterwordspacing

\bibitem{wang2025yolov9}
C.-Y. Wang, I.-H. Yeh, and H.-Y. Mark~Liao, ``{YOLOv9: Learning What You Want to Learn Using Programmable Gradient Information},'' in \emph{European Conference on Computer Vision (ECCV)}.\hskip 1em plus 0.5em minus 0.4em\relax Springer, 2025.

\bibitem{wang2024yolov10}
A.~Wang, H.~Chen, L.~Liu, K.~Chen, Z.~Lin, J.~Han, and G.~Ding, ``{YOLOv10: Real-Time End-to-End Object Detection},'' in \emph{Advances in Neural Information Processing Systems (NIPS)}, 2024.

\bibitem{yolo11_ultralytics}
\BIBentryALTinterwordspacing
G.~Jocher and J.~Qiu, ``{Ultralytics YOLO11},'' 2025. [Online]. Available: \url{https://github.com/ultralytics/ultralytics}
\BIBentrySTDinterwordspacing

\bibitem{tian2025yolov12}
Y.~Tian, Q.~Ye, and D.~Doermann, ``{YOLOv12: Attention-Centric Real-Time Object Detectors},'' \emph{arXiv preprint arXiv:2502.12524}, 2025.

\bibitem{zhao2024detrs}
Y.~Zhao, W.~Lv, S.~Xu, J.~Wei, G.~Wang, Q.~Dang, Y.~Liu, and J.~Chen, ``{DETRs Beat YOLOs on Real-time Object Detection},'' in \emph{Conference on Computer Vision and Pattern Recognition (CVPR)}, 2024.

\bibitem{lv2024rt}
W.~Lv, Y.~Zhao, Q.~Chang, K.~Huang, G.~Wang, and Y.~Liu, ``{RT-DETRv2: Improved Baseline with Bag-of-Freebies for Real-Time Detection Transformer},'' \emph{arXiv preprint arXiv:2407.17140}, 2024.

\bibitem{wang2025rt}
S.~Wang, C.~Xia, F.~Lv, and Y.~Shi, ``{RT-DETRv3: Real-time End-to-End Object Detection with Hierarchical Dense Positive Supervision},'' in \emph{Winter Conference on Applications of Computer Vision (WACV)}.\hskip 1em plus 0.5em minus 0.4em\relax IEEE, 2025.

\bibitem{mao20233d}
J.~Mao, S.~Shi, X.~Wang, and H.~Li, ``{3D Object Detection for Autonomous Driving: A Comprehensive Survey},'' \emph{International Journal of Computer Vision}, 2023.

\bibitem{brazil2019m3d}
G.~Brazil and X.~Liu, ``{M3D-RPN: Monocular 3D Region Proposal Network for Object Detection},'' in \emph{International Conference on Computer Vision (ICCV)}, 2019.

\bibitem{tian2019fcos}
Z.~Tian, C.~Shen, H.~Chen, and T.~He, ``{FCOS: Fully convolutional one-stage object detection},'' in \emph{International Conference on Computer Vision (ICCV)}, 2019.

\bibitem{mallot1991inverse}
H.~A. Mallot, H.~H. B{\"u}lthoff, J.~J. Little, and S.~Bohrer, ``Inverse perspective mapping simplifies optical flow computation and obstacle detection,'' \emph{Biological cybernetics}, vol.~64, no.~3, 1991.

\bibitem{luo2018mio}
Z.~Luo, F.~Branchaud-Charron, C.~Lemaire, J.~Konrad, S.~Li, A.~Mishra, A.~Achkar, J.~Eichel, and P.-M. Jodoin, ``{MIO-TCD: A New Benchmark Dataset for Vehicle Classification and Localization},'' \emph{IEEE Transactions on Image Processing}, vol.~27, no.~10, 2018.

\bibitem{wen2020ua}
L.~Wen, D.~Du, Z.~Cai, Z.~Lei, M.-C. Chang, H.~Qi, J.~Lim, M.-H. Yang, and S.~Lyu, ``{UA-DETRAC: A new benchmark and protocol for multi-object detection and tracking},'' \emph{Computer Vision and Image Understanding}, vol. 193, 2020.

\bibitem{guerrero2013iwinac}
R.~Guerrero-Gomez-Olmedo, R.~J. Lopez-Sastre, S.~Maldonado-Bascon, and A.~Fernandez-Caballero, ``{Vehicle Tracking by Simultaneous Detection and Viewpoint Estimation},'' in \emph{International Work-Conference on the Interplay Between Natural and Artificial Computation (IWINAC)}, 2013.

\bibitem{ye2023yolov7}
Z.~Ye, H.~Zhang, J.~Gu, and X.~Li, ``{YOLOv7-3D: A Monocular 3D Traffic Object Detection Method from a Roadside Perspective},'' \emph{Applied Sciences}, vol.~13, no.~20, 2023.

\bibitem{ye2022rope3d}
X.~Ye, M.~Shu, H.~Li, Y.~Shi, Y.~Li, G.~Wang, X.~Tan, and E.~Ding, ``{Rope3D: The Roadside Perception Dataset for Autonomous Driving and Monocular 3D Object Detection Task},'' in \emph{Conference on Computer Vision and Pattern Recognition (CVPR)}, 2022.

\bibitem{amirgholy2020optimal}
M.~Amirgholy, M.~Nourinejad, and H.~O. Gao, ``{Optimal traffic control at smart intersections: Automated network fundamental diagram},'' \emph{Transportation Research Part B: Methodological}, vol. 137, 2020.

\bibitem{zimmer2023tumtrafintersection}
W.~Zimmer, C.~Cre{\ss}, H.~T. Nguyen, and A.~C. Knoll, ``{TUMTraf Intersection Dataset: All You Need for Urban 3D Camera-LiDAR Roadside Perception},'' in \emph{International Conference on Intelligent Transportation Systems (ITSC)}.\hskip 1em plus 0.5em minus 0.4em\relax IEEE, 2023.

\bibitem{zipfl2020traffic}
M.~Zipfl, T.~Fleck, M.~R. Zofka, and J.~M. Z{\"o}llner, ``{From Traffic Sensor Data To Semantic Traffic Descriptions: The Test Area Autonomous Driving Baden-W{\"u}rttemberg Dataset (TAF-BW Dataset)},'' in \emph{International Conference on Intelligent Transportation Systems (ITSC)}.\hskip 1em plus 0.5em minus 0.4em\relax IEEE, 2020.

\bibitem{fleck2023semi}
T.~Fleck, M.~Zipfl, and J.~M. Z{\"o}llner, ``{Semi-Automatic Ground Truth Trajectory Estimation and Smoothing using Roadside Cameras},'' in \emph{International Conference on Intelligent Transportation Systems (ITSC)}.\hskip 1em plus 0.5em minus 0.4em\relax IEEE, 2023.

\bibitem{cvat}
{B. Sekachev, N. Manovich, M. Zhiltsov, A. Zhavoronkov, D. Kalinin, B. Hoff, et al.}, ``{CVAT},'' \url{https://github.com/cvat-ai/cvat}, 2025.

\bibitem{dosovitskiy2017carla}
A.~Dosovitskiy, G.~Ros, F.~Codevilla, A.~Lopez, and V.~Koltun, ``{CARLA: An Open Urban Driving Simulator},'' in \emph{Conference on Robot Learning}, 2017.

\bibitem{zheng2020distance}
Z.~Zheng, P.~Wang, W.~Liu, J.~Li, R.~Ye, and D.~Ren, ``{Distance-IoU loss: Faster and better learning for bounding box regression},'' in \emph{AAAI Conference on Artificial Intelligence}, 2020.

\bibitem{OpenImages}
A.~Kuznetsova, H.~Rom, N.~Alldrin, J.~Uijlings, I.~Krasin, J.~Pont-Tuset, S.~Kamali, S.~Popov, M.~Malloci, A.~Kolesnikov, T.~Duerig, and V.~Ferrari, ``{The Open Images Dataset V4: Unified image classification, object detection, and visual relationship detection at scale},'' \emph{IJCV}, 2020.

\bibitem{ruis2024improving}
F.~A. Ruis, A.~M. Liezenga, F.~G. Heslinga, L.~Ballan, T.~A. Eker, R.~J. den Hollander, M.~C. van Leeuwen, J.~Dijk, and W.~Huizinga, ``Improving object detector training on synthetic data by starting with a strong baseline methodology,'' in \emph{Synthetic Data for Artificial Intelligence and Machine Learning: Tools, Techniques, and Applications II}, 2024.

\end{thebibliography}
}

\end{document}